%% file: main.tex
\documentclass{ecai}
\usepackage{times}
\usepackage{graphicx}
\usepackage{amsmath}
\usepackage{amssymb}
\usepackage{bm}
\usepackage{mathtools}
\usepackage{multirow}
\usepackage{tikz}
\usepackage{algpseudocode,algorithm,algorithmicx}
\usetikzlibrary{shapes.geometric}
\usetikzlibrary{arrows,automata,shapes,calc,matrix,fit,arrows.meta}
\usetikzlibrary{positioning}
\usepackage{todonotes}
\usepackage{xcolor}

\DeclareMathOperator*{\argmax}{arg\,max}

\ecaisubmission   

\begin{document}

\title{Verification of Neural Networks:\\ Enhancing Scalability through Pruning}

\author{Dario Guidotti\thanks{University of Genoa, Italy, dario.guidotti@edu.unige.it} 
\and Francesco Leofante\thanks{Imperial College London, United Kingdom, f.leofante@imperial.ac.uk}
\and Luca Pulina\thanks{University of Sassari, Italy, lpulina@uniss.it}
\and Armando Tacchella\thanks{University of Genoa, Italy, armando.tacchella@unige.it}}

\maketitle
\bibliographystyle{ecai}

\begin{abstract}
  Verification of deep neural networks has witnessed a recent
  surge of interest, fueled by success stories in diverse domains 
  and by abreast concerns about safety and security in envisaged
  applications. Complexity and sheer size of such networks are
  challenging for automated formal verification techniques which, 
  on the other hand, could ease the adoption 
  of deep networks in safety- and security-critical contexts. 
  In this paper we focus on enabling state-of-the-art 
  verification tools to deal with neural networks of 
  some practical interest. We propose a new training pipeline
  based on network pruning with the goal of striking 
  a balance between maintaining accuracy and robustness while making the resulting networks amenable to
  formal analysis. The results of our experiments with a portfolio  
  of pruning algorithms and verification tools show that 
  our approach is successful for the kind of networks we consider
  and for some combinations of pruning and verification techniques, 
  thus bringing deep neural networks closer to the reach of
  formally-grounded methods.
\end{abstract}

\input{introduction}

\input{background}
\input{methodologies}
\input{casestudy}
\input{results}

\input{conclusions}


\bibliography{biblio}
\end{document}

%% file: introduction.tex
\section{Introduction}

Verification of neural networks is currently a hot topic involving
different areas of AI across machine learning, constraint programming,
heuristic search and automated reasoning. A recent extensive
survey~\cite{huang2018safety} cites more than 200 papers, most of
which have been published in the last few years, and more
contributions are coming out at an impressive rate --- see,
e.g.,~\cite{DBLP:conf/hybrid/DuttaCJST19,DBLP:conf/cav/KatzHIJLLSTWZDK19,DBLP:journals/corr/abs-1807-03571,DBLP:conf/aaai/NarodytskaKRSW18,DBLP:journals/corr/LomuscioM17,DBLP:conf/nips/WangPWYJ18} 
to cite only some. Most of the current
literature focuses on the verification of Deep Neural Networks (DNNs): 
while their application in various
domains~\cite{DBLP:journals/nature/LeCunBH15} have made them one of
the most popular machine-learned models to date, concerns about their
vulnerability to adversarial
perturbations~\cite{DBLP:journals/corr/SzegedyZSBEGF13,DBLP:journals/corr/GoodfellowSS14}
have been accompanying them since their initial adoption, to the point
of restraining their application in safety- and security-related
contexts. 

In this paper we present evidence that state-of-the-art verification
tools can deal with DNNs of some practical interest through
preprocessing based on statistical techniques. We propose a new
training pipeline that leverages network pruning --- a controlled reduction of
the size of neural networks to eliminate portions that are deemed not
crucial for their performances~\cite{DBLP:conf/icnn/SietsmaD88} --- to produce easier-to-verify networks.     
Our goal is to strike a balance between maintaining accuracy,
generalization power and robustness of our networks, while
making them amenable to formal analysis. In particular, we
consider two mainstream pruning techniques, namely neuron
pruning~\cite{DBLP:conf/icnn/SietsmaD88} and weight
pruning~\cite{DBLP:conf/nips/CunDS89}. Intuitively, the former
technique attempts to make DNNs leaner by removing neurons, and thus
severing all connections across different layers going through pruned
neurons; the latter acts on connections, removing
neurons only when all connections through a neuron are zeroed 
out. Both techniques make the network graph sparser, so the
key idea behind our approach is that, as long as sparse networks
retain acceptable performances, their verification
might be easier than that of their dense counterparts~\cite{DBLP:conf/iclr/XiaoTSM19}, enabling deployment in
contexts where formal guarantees are of the essence, and smaller
networks are a welcome bonus.

To put our idea to the test, we consider fully-connected DNNs trained
on well-known datasets --- namely MNIST~\cite{lecun1998gradient}
and Fashion MNIST~\cite{DBLP:journals/corr/abs-1708-07747}
--- and a portfolio of verification
tools to assess the gain brought by pruning to the verification
phase. Among the plethora of verification methods
available~\cite{huang2018safety}, we focus on the kind surveyed
in~\cite{DBLP:journals/corr/abs-1805-09938} characterized by a ``push  
button'' approach in which, given input pre-conditions and output
post-conditions stated as constraints in some formal logic, a tool is
asked to answer an entailment query in such logic, i.e., whether the
post-conditions are satisfied given that pre-conditions hold. 
We further focus on tools that are distributed as
system prototypes, and assemble our portfolio considering diverse approaches
within the class we focus on. In particular, we consider
Marabou~\cite{DBLP:conf/cav/KatzHIJLLSTWZDK19},
ERAN~\cite{DBLP:conf/iclr/SinghGPV19} 
and MIPVerify~\cite{DBLP:conf/iclr/TjengXT19}.
Marabou transforms the entailment query into constraint satisfaction
problems solved with a dedicated satisfiabilty modulo theory (SMT)
solver. ERAN tackles the certification of neural networks against
adversarial perturbations by combining over-approximation methods with
precise mixed integer linear programming (MILP). Finally, MIPVerify is
a robustness analyzer completely based on a compilation to MILP.  

Through the combination of weight and neuron pruning and our selection
of verification tools we can observe the following results:
\begin{itemize}
\item Without network pruning, the verification of relatively small
  fully connected networks --- about 50K connections on 896 neurons
  --- is mostly out of the reach of complete verification tools 
  within a timeout of 10 CPU minutes.
\item Neuron pruning is a stronger enabler than weight pruning; all
  tools benefit from it, and for some combinations of parameters and
  verification tools, neuron pruning enables the verification of all
  the network instances that we consider within the timeout.
\item  Weight pruning appears to be only effective for small networks
and high magnitude of pruning (\textit{e.g.}, $90\%$ of the total weights). In particular ERAN, both in its complete
and incomplete version seems to be able to reap more benefits of weight pruning than MIPVerify and Marabou.
\end{itemize}

While our findings are related to fully-connected networks trained
and tested on specific datasets, we believe that similar advantages
can be obtained considering more complex feed-forward architectures
including, e.g., convolutional or residual layers, and other
standard datasets which will be the subject of our continuing research.

The rest of the paper is structured as follows. In
Section~\ref{sec:background} we introduce some basic notation and
definitions to be used throughout the paper. In
Section~\ref{sec:methodologies} we introduce our approach, detailing
the algorithms for different kinds of pruning and other modifications
required to make networks amenable to formal analysis. In
Section~\ref{sec:casestudy} we describe the datasets on which our
experimental analysis is based, and the specific conditions that we
asked the verification tools to check. In Section~\ref{sec:experiments} we
present the setup and the results of our experimental analysis. We
conclude the paper in Section~\ref{sec:conclusions} with some final
remarks and our agenda for future research.

%% file: background.tex
\section{Background}\label{sec:background}
A \emph{neural network} is a system of interconnected computing
units called \emph{neurons}.  
In \emph{fully connected feed-forward} networks, neurons are arranged in
disjoint layers, with each layer being fully connected only with
the next one, and without connection between neurons in the same
layer. Given a feed-forward neural network $N$ with $n$ layers, we
denote the $i$-th layer of $N$ as $\mathbf{h}^{(i)}$. We call a layer
without incoming connections \emph{input} layer $\mathbf{h}^{(1)}$,
a layer without outgoing connections \emph{output} layer
$\mathbf{h}^{(n)}$, while all other layers are referred to as 
\emph{hidden} layers . 
Each hidden layer performs specific transformations on the inputs
it receives. In this work we consider hidden layers that
make use of \emph{linear}  and \emph{batch normalization}
modules. Given an input vector $\mathbf{x} $, a linear module computes
a linear combination of its values as follows:   
\begin{align}
\begin{split}
&\mathbf{L}^{(i)} = \mathbf{W}^{(i)} \cdot \mathbf{x} + \mathbf{b}^{(i)}
\label{eq:dnn-ext}
\end{split}
\end{align}
where $\mathbf{W}^{(i)} $ is the matrix of weights and 
$\mathbf{b}^{(i)} $ is the vector of
the biases associated with the linear module in the $i$-th layer and $\mathbf{L}^{(i)} $ is the corresponding output. Entries of both $\mathbf{W}^{(i)}$ and $\mathbf{b}^{(i)}$ are learned parameters.
In our target architectures, each linear module is followed by a batch normalization module. This is done
to address the so-called \emph{internal covariate shift} problem,
i.e., the change of the distribution of each layer's input during
training~\cite{DBLP:conf/icml/IoffeS15}.  
The mathematical formulation of batch normalization layers can be expressed as 
\begin{equation}
\mathbf{BN}^{(i)}  = \frac{\bm{\gamma}^{(i)}}{\sqrt[2]{\bm{\sigma}^{(i)} + \epsilon^{(i)}}} \odot (\mathbf{L}^{(i)} - \bm{\mu}^{(i)}) + \bm{\beta}^{(i)}
\label{eq:batch-norm}
\end{equation}
All the operators in this equation are element-wise operators: in 
particular $\odot$ and the fractional symbol represent respectively
the Hadamard 
product and division. 
$\mathbf{BN}^{(i)}$ and $\mathbf{L}^{(i)}$ are the 
output and the input vectors of the module, respectively.
$\bm{\gamma}^{(i)}$, $\bm{\mu}^{(i)}$, $\bm{\sigma}^{(i)}$,
$\bm{\beta}^{(i)}$ are vectors, whereas $\epsilon^{(i)}$
is a scalar value. These are learned parameters of the batch normalization
layer. In particular $\bm{\mu}^{(i)}$ and  $\bm{\sigma}^{(i)}$ are
the estimated mean and variance of the inputs computed during training.\\
Finally, the output of hidden layer $i$ is computed as $\mathbf{h}^{(i)} = \Phi^{(i)} (\mathbf{BN}^{(i)})$,
where $\Phi^{(i)}$ is the \emph{activation function} associated to
the neurons in the layer. We consider only networks utilizing
\emph{Rectified Linear Unit (ReLU)} activation functions, i.e.,  $\Phi^{(i)} = \max(0, \mathbf{BN}^{(i)})$.
Given an input vector $\mathbf{x}$, the network $N$ computes an output vector $\mathbf{y}$ by means of the
following computations
\begin{align}
\begin{split}
& \mathbf{h}^{(1)} = \mathbf{x}\\
& \mathbf{h}^{(i)} = \Phi^{(i)} (\mathbf{BN}^{(i)}(\mathbf{L}^{(i)} (\mathbf{h}^{(i-1)}))) \quad i=2,\ldots,n-1\\
& \mathbf{y} = \mathbf{h}^{(n)} = \mathbf{L}(\mathbf{h}^{(n-1)})
\label{eq:dnn}
\end{split}
\end{align}

A neural network can be considered as a non-linear function
$f_\textbf{w}: \mathcal{X} \rightarrow \mathcal{Y}$, where
$\mathcal{X}$ is the input space of the network, $\mathcal{Y}$ is
the output space and $\mathbf{w}$ is the vector representing the weights of all the connections. We consider neural network applied to classification 
of $d$-dimensional vectors of real numbers, i.e., 
$\mathcal{X} \subseteq \mathbb{R}^d$ and $\mathcal{Y} \subseteq \mathbb{R}^m$, where $d$ is the dimension of the input vector and $m$ is the dimension of the output vector and thus also the number of possible classes of interest.
We assume that given an input sample $\mathbf{x}$ the output vector
$f_\mathbf{w}(\mathbf{x})$ contains the likelihood that $\mathbf{x}$
belongs to one of the $m$ classes. The specific class can be computed
as
$$
\argmax\limits_{i \in \{1, \ldots, m\}} (f_{\mathbf{w}}(\mathbf{x}))_c
$$
where $(f_{\mathbf{w}}(\mathbf{x}))_c$ denotes the $c$-th element of
$f_\mathbf{w}$.
Training of (deep) neural networks poses substantial computational challenges
since for state-of-the-art models the size of $\mathbf{w}$ can be in the order of millions.
As in any machine learning task, training must select weights
so as to maximize the likelihood that the networks responds correctly,
i.e., if the input $\mathbf{x}$ is of class $k$, the chance of
\emph{misclassification} should be as small as possible, where
misclassification occurs whenever the following holds  
$$
\argmax\limits_{c \in \{1, \ldots, m\}} (f_{\mathbf{w}}(\mathbf{x}))_c \neq k
$$
Training can be achieved through minimization of some kind of
\emph{loss function} whose value is low when the chance of
misclassification is also low. While there are many different kinds of
loss functions, in general they are structured in the following way:     
\begin{equation}
J(\mathbf{w}) = \frac{1}{n} \sum_{k=0}^n Err(y_k, \argmax\limits_{i \in \{1, \ldots, m\}} (f_{\mathbf{w}}(\mathbf{x}_k))_i) + \lambda \cdot Reg(\mathbf{w})
\end{equation}
where $n$ is the number of training pairs $(\mathbf{x}_k, y_k)$,
$y_k$ is the correct class label of $\mathbf{x}_k$, 
$Err$ represents the loss caused by misclassification, 
$Reg$ is a \emph{regularization} function, and $\lambda$ is the
parameter controlling the effect of $Reg$ on $J$.
The regularization function is needed to avoid \emph{overfitting},
i.e., high variance of the training results with respect to the
training data. The regularization function usually penalizes models with high complexity by smoothing out sharp variations induced in the trained network by the $Err$ function. A common
regularization function is, for example, the L2 norm: 
\begin{equation}
Reg(\mathbf{w}) = \frac{1}{2n} ||\mathbf{w}||_2
\end{equation}

%% file: methodologies.tex
\section{Pruning for verification: training pipeline}
\label{sec:methodologies}

In this Section we present the core algorithms of our approach
based on network pruning (NP) and weight pruning (WP).
Concerning NP, its  aim is to remove neurons from the network,
together with the corresponding input and  
output connections. Algorithm~\ref{alg:np} reports our approach
based on NP techniques that we adapted from those proposed in~\cite{DBLP:conf/iccv/LiuLSHYZ17} for convolutional neural networks.   
Looking at the pseudocode in Algorithm~\ref{alg:np}, we can see that
the procedure takes as input (Line 1) the NN to train
($f_{\mathbf{w}}$), the sparsity rate ($\rho$) --- which indicates the
percentage of neurons to prune --- and a regularizer ($\lambda$);
the pruned network is returned as output (Line 8).
The first step of the algorithm consists in training the network (Line
2) using a loss function with regularization element ($\lambda$)
which encourages to nullify the parameters $\bm{\gamma}^{(i)}$ of the
batch normalization layers. The regularization term we consider in
\textsc{sparseTraining} can be formalized as $\lambda \cdot
\sum_{i=2}^{n} ||\bm{\gamma}^{(i)}||_1$, referring to the network
topology presented in Section \ref{sec:background}. Each of these
parameters identifies a neuron of the previous linear layer.
In Line 3 and 4, we extract the weight parameters (\textit{i.e.},
$\bm{\gamma}_i$) of the batch normalization layers and we sort them in
ascending order. In Line 5, we select as threshold the weight with
index equals to the number of neurons  we want to prune (given by the
Sparsity Rate $\rho$ multiplied for the number of weights): in this
way the number of weights below the threshold corresponds exactly to
the number of neuron we want to prune.
\textsc{pruneNeurons} (Line 6) applies neuron pruning by eliminating
all the neurons corresponding to the weights of the batch
normalization layers which are below the threshold computed in Line 5.
Finally, the pruned network is trained again in order to remedy to the
loss of accuracy caused by pruning (\textsc{fineTune}, Line 7).

\begin{algorithm}[t!]
	\caption{Neuron Pruning}
	\label{alg:np}
	\begin{algorithmic}[1]
		\Function{neuronPruning}{$f_{\mathbf{w}}$, $\rho$, $\lambda$}
		\State $f_{\tilde{\mathbf{w}}} \gets $ \Call{sparseTraining}{$f_{\mathbf{w}}$, $\lambda$}
		\State $weights \gets $ \Call{extractWeightsBatchnorm}{$f_{\tilde{\mathbf{w}}}$}
		\State $incWeights \gets $ \Call{incOrder}{$weights$}\;
		\State $threshold \gets  incWeights[\rho \cdot $ \Call{len}{$weights$}$]$
		\State $\hat{f}_{\tilde{\mathbf{w}}} \gets $ \Call{pruneNeurons}{$f_{\tilde{\mathbf{w}}}$, $threshold$}
		\State $\hat{f}_{\widehat{\mathbf{w}}} \gets $ \Call{fineTune}{$\hat{f}_{\tilde{\mathbf{w}}}$}
		\State \Return $\hat{f}_{\widehat{\mathbf{w}}}$
		\EndFunction
	\end{algorithmic}
\end{algorithm}

\begin{algorithm}[t!]
\caption{Weight Pruning}
\label{alg:wp}
	\begin{algorithmic}[1]
          \Function{weightPruning}{$f_{\mathbf{w}}$, $\rho$}
          \State $f_{\tilde{\mathbf{w}}} \gets $ \Call{Training}{$f_{\mathbf{w}}$}
          \State $weights \gets $ \Call{extractWeightsLinear}{$f_{\tilde{\mathbf{w}}}$}
          \State $incWeights \gets $ \Call{incOrder}{$weights$}
          \State $threshold \gets  incWeights[\rho \cdot $ \Call{len}{$weights$}$]$
          \State $f_{\tilde{\mathbf{w}}} \gets $ \Call{pruneWeights}{$f_{\mathbf{w}}$, $threshold$}
          \State $f_{\widehat{\mathbf{w}}} \gets $ \Call{fineTune}{$f_{\tilde{\mathbf{w}}}$}
          \State \Return $f_{\widehat{\mathbf{w}}}$
 	  \EndFunction
        \end{algorithmic}
\end{algorithm}

Considering WP, the general aim of this technique is to
eliminate weights, i.e., set them to $0$. Also in this case, weights
are pruned if their values after training are below a
user-specified threshold. Algorithm~\ref{alg:wp} shows the pseudocode
of our approach inspired by~\cite{DBLP:journals/corr/HanPTD15}. 
In Line 2 we train a NN using a standard training procedure. In lines 3 and 4, the weights of the linear layers are extracted from
the original network $f_{\tilde{\mathbf{w}}}$ and then are sorted in ascending order. 
In Line 5 we select the threshold from ordered weights based on the
\emph{Sparsity Rate} parameter $\rho$, in the same way as we do in the
NP algorithm. Then, \textsc{pruneWeights} (Line 6) sets to $0$ all the
weights of the network of interest which are smaller than the
threshold, hence obtaining the pruned network
$f_{\widehat{\mathbf{w}}}$. 
Analogously to what we do for NP, at the end of the procedure (Line 7)
we \emph{fine tune} the pruned network. 

\paragraph{Encoding batch normalization layers as linear layers}
Not all the verification tools accept networks with batch
normalization layers. For this reason, our final models need to be
devoid of such modules, but training and neuron pruning require those
layers to function properly. To overcome this hurdle, we propose a
post-processing technique to merge together a batch normalization layer and the 
previous linear layer in a new linear layer. Merging is
performed after training and pruning. To see how merging is
performed, let us consider again the expressions for a generic linear
layer and its subsequent batch normalization layer:
\begin{align}
\begin{split}
&\mathbf{L}^{(i)} = \mathbf{W}^{(i )} \cdot \mathbf{h}^{(i - 1)} + \mathbf{b}^{(i)}\\
&\mathbf{BN}^{(i)} = \frac{\bm{\gamma}^{(i)}}{\sqrt[2]{\bm{\sigma}^{(i)} + \epsilon^{(i)}}} \odot (\mathbf{L}^{(i)} - \bm{\mu}^{(i)}) + \bm{\beta}^{(i)}
\label{eq:norm-fusion1}
\end{split}
\end{align}
By substituting the value of $\mathbf{L}^{(i)}$ in the second equation with the 
value given by the first one we obtain:
\begin{align}
\begin{split}
\mathbf{BN}^{(i)} = &\ \mathbb{I} \cdot \frac{\bm{\gamma}^{(i)}}{\sqrt[2]{\bm{\sigma}^{(i)} + \epsilon^{(i)}}} \cdot \mathbf{W}^{(i)} \cdot \mathbf{h}^{(i - 1)} +\\ 
&+ \frac{\bm{\gamma}^{(i)}}{\sqrt[2]{\bm{\sigma}^{(i)} + \epsilon^{(i)}}} \odot (\mathbf{b}^{(i )} - \bm{\mu}^{(i)}) + \bm{\beta}^{(i)}
\label{eq:norm-fusion2}
\end{split}
\end{align}
where $\mathbb{I}$ represent the identity matrix with dimension $d
\times d$, and all the other symbols have the same meaning of the
corresponding ones in equations \ref{eq:dnn-ext} and \ref{eq:batch-norm}.
Therefore we can express the operations performed by the two layers by means of a new linear layer defined with the following parameters
\begin{align}
\begin{split}
&\mathbf{W}^{(i)} := \mathbb{I} \cdot \frac{\bm{\gamma}^{(i)}}{\sqrt[2]{\bm{\sigma}^{(i)} + \epsilon^{(i)}}} \cdot \mathbf{W}^{(i )} \\
&\mathbf{b}^{(i)} := \frac{\bm{\gamma}^{(i)}}{\sqrt[2]{\bm{\sigma}^{(i)} + \epsilon^{(i)}}} \odot (\mathbf{b}^{(i )} - \bm{\mu}^{(i)}) + \bm{\beta}^{(i)}
\label{eq:norm-fusion3}
\end{split}
\end{align}

%% file: casestudy.tex
\section{Case studies}
\label{sec:casestudy}

In our experimental analysis we consider networks trained on 
MNIST~\cite{lecun1998gradient} and Fashion MNIST
(FMNIST)~\cite{DBLP:journals/corr/abs-1708-07747} datasets. Both
datasets consist of 70000 grayscale images of $28 \times 28$ pixels
divided up in 10 different classes, each class represented by
the same number of images. The datasets are both divided in a training
set of 60000 images and a test set of 10000 images. The difference
between the two datasets is that MNIST images represent handwritten
digits, whereas FMNIST images represent fashion articles. While
the two datasets are similar, it turns out that training neural
classifiers for FMNIST is harder than solving the same task for MNIST,   
as testified in~\cite{DBLP:journals/corr/abs-1708-07747}.  

To perform verification of the trained networks, we consider the
absence of $L_\infty$-bounded adversarial examples as our main
requirement. We recall that the \emph{infinity norm},
denoted as $L_\infty$, is defined for any vector $\mathbf{x} = (x_1,
\ldots, x_d)$ as  
$$
||\mathbf{x}||_\infty = \max(|x_1|, \ldots, |x_d|).
$$
Besides simplicity and generality --- the existence of
$L_\infty$-bounded adversarial examples can be easily stated in all
the verification tools we consider --- the 
absence or timely detection of such adversarial examples is 
relevant for practical
applications~\cite{DBLP:journals/corr/SzegedyZSBEGF13}.    
We can formalize absence of $L_\infty$-bounded adversarial examples
for networks performing image classification tasks in the following
way. Let be $f_{\mathbf{w}} : \mathcal{X} \to \mathcal{Y}$ be a neural
network for image classification as defined in
Section~\ref{sec:background}.
Given an input image $\hat{\mathbf{x}} \in \mathcal{X}$
and a bound $\varepsilon$ on the infinity norm, we can express a
\emph{targeted $L_\infty$-bounded adversarial example of class} $k \in \{1,
\ldots, m\}$ as an image $\mathbf{x}$ such that 
\begin{equation}
||\mathbf{x} - \hat{\mathbf{x}}||_\infty < \varepsilon \:\Rightarrow\:
\argmax\limits_{c \in \{1, \ldots, m\}}  (f_{\mathbf{w}}(\mathbf{x}))_c = k
\label{eq:tar-adv}
\end{equation}
where ``$\Rightarrow$'' denotes logical implication and $k$ is the
class assigned by the network to the adversarial 
$\mathbf{x}$. The corresponding \emph{untargeted $L_\infty-bounded$
  adversarial example} 
is an image $\mathbf{x}$ such that 
\begin{equation}
||\mathbf{x} - \hat{\mathbf{x}}||_\infty < \varepsilon \:\Rightarrow\:
\argmax\limits_{c \in \{1, \ldots, m\}} (f_{\mathbf{w}}(\mathbf{x}))_c \neq j
\label{eq:untar-adv}
\end{equation}
where $j$ is the correct label for the image $\hat{\mathbf{x}}$. In
the following we speak of targeted and untargeted adversarial examples
meaning the $L_\infty$-bounded versions defined above, and we describe
in detail how we encoded the search for adversarial examples in the
tools that we consider.

The encoding of targeted adversarial examples for Marabou is
straightforward. Considering a network
$f_{\mathbf{w}} : \mathcal{X} \to \mathcal{Y}$ with 
$\mathcal{X} \subseteq \mathbb{R}^d$
and $\mathcal{Y} \subseteq \mathbb{R}^m$, given 
an input image $\hat{\mathbf{x}} \in \mathcal{X}$, 
the target adversarial class $k$ and the bound on the infinity norm
$\epsilon$ we can encode (\ref{eq:tar-adv})
as: 
\begin{align}
  \begin{array}{ll}
\hat{x}_i - \varepsilon \leq x_i \leq
\hat{x}_i + \varepsilon & \mbox{for } 0 \leq i \leq d\\
y_k \geq y_j & \mbox{for } 0 \leq j \leq m, \quad j \neq k
  \end{array}
\label{eq:marabou-advsearch}
\end{align}
where $\hat{x}_i$ denotes the $i$-th component of vector $\hat{\mathbf{x}}$,
$x_i$ with $i \in \{1, \ldots, d\}$ and $y_j$ with $j \in \{1, \ldots, m\}$
are $d + m$ search variables, tagged as input and output
variables, respectively: Marabou connects input and output variables
through the encoding of $f_{\mathbf{w}}$ performed automatically ---
we refer to~\cite{DBLP:conf/cav/KatzHIJLLSTWZDK19} for more
details.
The encoding of the untargeted adversarial search is
currently beyond our capability because Marabou does not offer an easy
way to encode disjunction of constraints like $\bigvee_k (y_k \geq
y_j)$ that would be required to express condition
(\ref{eq:untar-adv}). For this reason, we consider only targeted
adversarial search in the experiments with Marabou. 

ERAN provides untargeted adversarial example search only.
The encoding of the property and the network is performed automatically
by the tool, given the network $f_{\mathbf{w}}$ and an image
$\hat{\mathbf{x}}$. In particular, ERAN uses an abstract interpretation
approach in order to compute a symbolic overapproximation
$\overline{\mathcal{Y}}$ of the concrete outputs of the neural network
$f_{\mathbf{w}}$, given a symbolic overapproximation $\overline{\mathcal{X}}$
of the set of possible inputs. The overapproximation is such that
$f_{\mathbf{w}}(\mathbf{x}) \in \overline{\mathcal{Y}}$ as long as
$\mathbf{x} \in \overline{\mathcal{X}}$, i.e., all concrete outputs
are accounted for, as long as the concrete input is contained within
the boundaries of the input overapproximation.
In the case of $L_\infty$ bounded adversarial search we have that
$\overline{\mathcal{X}}$ contains all the vectors $\mathbf{x}$ such
that $||\mathbf{x} - \mathbf{x}||_\infty \leq \varepsilon$: the
corresponding output approximation $\overline{\mathcal{Y}}$ may
contain values for which the likelihood 
of the correct label is smaller than some other label. ERAN features
two versions: an incomplete one, where the output approximation is
inspected to check whether the network is safe with respect to adversarial;
a complete version, where ERAN takes one more step and calls a MILP
solver in order to find and output a concrete adversarial example.
More details about ERAN can be found in~\cite{DBLP:conf/iclr/SinghGPV19}.

MIPVerify can search for both targeted and untargeted adversarial
examples and it also automatically generates its internal encoding
given the network of interest and a starting input image. In
particular MIPVerify encodes the search of the 
untargeted adversarial example as the following MILP problem: 
\begin{equation}
  \begin{array}{ll}
    \multicolumn{2}{l}{\min_{\mathbf{x}} \varepsilon}\\
    \multicolumn{2}{l}{\argmax\limits_{c \in \{1, \ldots, m\}}(f_{\mathbf{w}}(\mathbf{x}))_c \neq j}\\
    x_i \leq \hat{x}_i + \varepsilon & \mbox{for } 0 \leq i \leq d\\
    x_i \geq \hat{x}_i - \varepsilon & \mbox{for } 0 \leq i \leq d\\
    \label{eq:mip-advsearch}
  \end{array}
\end{equation}
As can be seen from (\ref{eq:mip-advsearch}), MIPVerify searches
for the adversarial example which is closer to the original
image. The encoding of the network $f_{\mathbf{w}}$ is based on the
formulation of piece-wise linear functions as conjunctions of MILP
constraints. For an in-depth explanation of the complete encoding we refer
to~\cite{DBLP:conf/iclr/TjengXT19}. 

%% file: results.tex
\section{Experiments}\label{sec:experiments}
\subsection{Setup}\label{subsec:exp-setup}

We consider two different baseline architectures: both of them consists of an input
layer with $784$ neurons and an output layer with $10$ neurons. They both present three fully connected hidden layers with
$64$, $32$, $16$ neurons and $128$, $64$, $32$ neurons respectively. We will identify them as NET1 and NET2 respectively. \
Each hidden layer is followed by a batch
normalization layer and activation functions are Rectified Linear
Units (ReLUs). Networks are trained using an SGD optimizer with
Nesterov momentum~\cite{DBLP:conf/icml/SutskeverMDH13}.  
The learning parameters are:
\begin{itemize}
   \item \emph{learning rate} $= 0.1$, defining how quickly the model
     replaces the concepts it has learned with new ones, i.e., 
     controlling how much the optimizer can modify the network weights at
     each iteration;
   \item \emph{momentum} $= 0.9$,
     explained in~\cite{DBLP:conf/icml/SutskeverMDH13}, to control the
     gradient evolution in the attempt to escape local minima;  
   \item \emph{weight decay} $= 10^{-4}$, i.e., the magnitude of the
     regularization; 
   \item \emph{batch size} $= 64$, defining the number of samples
     considered before updating the model. 
\end{itemize}
We also use a learning rate scheduler which multiplies the learning
rate with a factor of $0.1$ whenever the loss stops decreasing for
more than $3$ consecutive epochs. The learning proceeds for $100$
epochs unless the loss stops decreasing for more than $10$ consecutive
epochs, in which case it terminates. 
All the learning and pruning algorithm we present are implemented
using the learning framework PyTorch~\cite{paszke2017automatic}
with GPU-intensive computation running on top of Google Colaboratory service.\\

\begin{table}
\centering
\begin{tabular}{|c|c|c|c|}
\hline
\textbf{Param} & \textbf{Neuron SR} & \textbf{Weight SR} & \textbf{Regularizer} \\
\hline
\textbf{SET1} & $\rho_{np} = 0.75 / 0.8$ & $\rho_{wp} = 0.9$ & $\lambda = 10^{-1}$ \\
\textbf{SET2} & $\rho_{np} = 0.5$ & $\rho_{wp} = 0.7$ & $\lambda = 10^{-4}$ \\
\textbf{SET3} & $\rho_{np} = 0.25$ & $\rho_{wp} = 0.5$ & $\lambda = 0.5 \cdot 10^{-4}$ \\
\hline
\end{tabular}
\caption{Sets of parameters considered in our
  experiment. \textbf{Param} represents the identifiers of the
  different parameter sets. \textbf{Neuron SR} and \textbf{Weight SR}
  represent the sparsity rate (SR) for neuron pruning and weight
  pruning, respectively. \textbf{Regularizer} represent the regularizer value for sparse training in neuron pruning.
  Neuron SR for SET3 is $0.75$ for NET1 and $0.8$ for NET2.}
\label{tab:par-set}
\end{table}

Table \ref{tab:par-set} shows the three different sets of pruning
parameters considered in our experiments. Neuron SR and Weight SR are
the sparsity rates supplied to the neuron pruning and weight pruning
procedures. We test three different sets of parameter to analyze how the
performances of the verification tools change with respect to
different amounts of pruning, with SET1 being the most aggressive
between the three.  Our implementation of  weight pruning
is based on the code described
in~\cite{DBLP:conf/iclr/LiuSZHD19}, and our implementation of 
neuron pruning is based on the code described
in~\cite{DBLP:conf/iccv/LiuLSHYZ17}. The setup presented is used for
the networks trained on MNIST as well as FMNIST. 

For each baseline network and dataset we
test the following: 
\begin{itemize}
\item \textit{Baseline Network}: this is the network trained using the standard training method without the regularizer $\lambda$.
\item \textit{Sparse Network}: this is the network trained using the sparse training method with the regularizer $\lambda$.
\item \textit{NP Network}: this is the network trained with Algorithm~\ref{alg:np} presented in Section~\ref{sec:methodologies}.
\item \textit{WP Network}: this is the network trained with Algorithm~\ref{alg:wp} presented in Section~\ref{sec:methodologies}.
\end{itemize}
For each such network we compute the accuracy with respect to the
original test set and the robustness using the tool
Foolbox~\cite{DBLP:journals/corr/RauberBB17}. In particular we
consider the $L_\infty$ bounded adversarial attack Gradient Sign
Method (GSM)~\cite{DBLP:journals/corr/GoodfellowSS14} in order to compute
adversarial examples.
 We use this method to assemble twenty images for which an adversarial
example can be found or can not be found by Foolbox on all the variant of the network of
interest (\textit{i.e.}, the adversarial is found on the same image for Baseline, Sparse, NP and WP
networks or it is not found for none of them.) --- notice that this does not mean
that an adversarial does not exist, because GSM is sound but not
complete with respect to the set of all bounded adversarial examples
of a given network.
The images resulting from this process are
the ones fed to verification tools in order to test their performances.
For all our experiments, we consider $\varepsilon = 0.1$ as the bound
on the infinity norm. The tools are tested with a timeout of $600$
seconds for each adversarial example, running ERAN in its complete
verification version.

One last remark about the setup concerns how we run the tools on
different problems. In particular, for Marabou we test the
images for which an adversarial example was found by searching for an adversarial example of the same class found by
Foolbox, whereas we test only one image for which the adversarial was not found, but considering
all the possible classes, one by one, for the adversarial
example. Because of this, we consider 19 instances of the
adversarial search, as opposed to the 20 instances considered for the  
other tools. For ERAN and MIPVerify we consider untargeted adversarial
example search only.

%
%
%
%
\subsection{Results}\label{subsec:exp-results}

\begin{table}
\centering
\begin{tabular}{|c|c|c|c|c|}
\hline
\multicolumn{5}{|c|}{\textbf{MNIST}}\\
\hline
\textbf{Base} & \textbf{Param} & \textbf{Network} & \textbf{Accuracy} & \textbf{Robustness} \\
\hline
\multirow{10}{4em}{\textbf{NET1}} & \multirow{1}{4em}{} & Baseline & $0.982$ & $0.322$ \\
\cline{2-5}
&\multirow{3}{4em}{\textbf{SET1}} & Sparse & $0.961$ & $0.200$ \\
& & WP & $0.982$ & $0.346$ \\
& & NP & $0.951$ & $0.170$ \\
\cline{2-5}
& \multirow{3}{4em}{\textbf{SET2}} & Sparse & $0.984$ & $0.353$ \\
& & WP & $0.983$ & $0.346$ \\
& & NP & $0.975$ & $0.204$ \\
\cline{2-5}
& \multirow{3}{4em}{\textbf{SET3}} & Sparse & $0.982$ & $0.276$ \\
& & WP & $0.984$ & $0.357$ \\
& & NP & $0.982$ & $0.264$ \\
\hline
\multirow{10}{4em}{\textbf{NET2}} & \multirow{1}{4em}{} & Baseline & $0.985$ & $0.360$ \\
\cline{2-5}
&\multirow{3}{4em}{\textbf{SET1}} & Sparse & $0.963$ & $0.151$ \\
& & WP & $0.985$ & $0.405$ \\
& & NP & $0.963$ & $0.151$ \\
\cline{2-5}
& \multirow{3}{4em}{\textbf{SET2}} & Sparse & $0.987$ & $0.353$ \\
& & WP & $0.987$ & $0.405$ \\
& & NP & $0.984$ & $0.309$ \\
\cline{2-5}
& \multirow{3}{4em}{\textbf{SET3}} & Sparse & $0.987$ & $0.363$ \\
& & WP & $0.987$ & $0.418$ \\
& & NP & $0.986$ & $0.367$ \\
\hline
\end{tabular}
\caption{Accuracy and Robustness of the networks of interest for the dataset MNIST. See Table \ref{tab:par-set} for the meaning of \textbf{Param}. The column \textbf{Network} represent the different networks we have considered in our experiments. The column \textbf{Accuracy} represent the accuracy of the networks computed on the test set. \textbf{Robustness} represent the robustness of the networks computed using Foolbox. Base represent which baseline architecture was used.}
\label{tab:MNIST-acc-rob}
\end{table}
\begin{table}
\centering
\begin{tabular}{|c|c|c|c|c|}
\hline
\multicolumn{5}{|c|}{\textbf{FMNIST}}\\
\hline
\textbf{Base} & \textbf{Param} & \textbf{Network} & \textbf{Accuracy} & \textbf{Robustness} \\
\hline
\multirow{10}{4em}{\textbf{NET1}} & \multirow{1}{4em}{} & Baseline & $0.894$ & $0.051$ \\
\cline{2-5}
& \multirow{3}{4em}{\textbf{SET1}} & Sparse & $0.855$ & $0.030$ \\
& & WP & $0.887$ & $0.023$ \\
& & NP & $0.854$ & $0.030$ \\
\cline{2-5}
& \multirow{3}{4em}{\textbf{SET2}} & Sparse & $0.894$ & $0.052$ \\
& & WP & $0.895$ & $0.035$ \\
& & NP & $0.885$ & $0.052$ \\
\cline{2-5}
& \multirow{3}{4em}{\textbf{SET3}} & Sparse & $0.894$ & $0.038$ \\
& & WP & $0.894$ & $0.047$ \\
& & NP & $0.893$ & $0.052$ \\
\hline
\multirow{10}{4em}{\textbf{NET2}} & \multirow{1}{4em}{} & Baseline & $0.895$ & $0.039$ \\
\cline{2-5}
& \multirow{3}{4em}{\textbf{SET1}} & Sparse & $0.856$ & $0.045$ \\
& & WP & $0.895$ & $0.025$ \\
& & NP & $0.871$ & $0.041$ \\
\cline{2-5}
& \multirow{3}{4em}{\textbf{SET2}} & Sparse & $0.899$ & $0.046$ \\
& & WP & $0.896$ & $0.031$ \\
& & NP & $0.895$ & $0.045$ \\
\cline{2-5}
& \multirow{3}{4em}{\textbf{SET3}} & Sparse & $0.901$ & $0.054$ \\
& & WP & $0.899$ & $0.039$ \\
& & NP & $0.899$ & $0.074$ \\
\hline
\end{tabular}
\caption{Accuracy and Robustness of the networks of interest for the dataset FMNIST. See Table \ref{tab:MNIST-acc-rob} for the meaning of the columns.}
\label{tab:FMNIST-acc-rob}
\end{table}
%
%

\begin{table}
\centering
\begin{tabular}{|c|c|c|c|c|}
\hline
\textbf{Base} & \textbf{Param} & \textbf{Network} & \textbf{\# HL Neurons} & \textbf{\# Weights} \\
\hline
\multirow{10}{4em}{\textbf{NET1}} & \multirow{1}{4em}{} & Baseline & $112$ & $52896$ \\
\cline{2-5}
& \multirow{3}{4em}{\textbf{SET1}} & Sparse & $112$ & $52896$ \\
& & WP & $112$ & $5290$ \\
& & NP & $27$ & $9624$ \\
\cline{2-5}
& \multirow{3}{4em}{\textbf{SET2}} & Sparse & $112$ & $52896$ \\
& & WP & $112$ & $15869$ \\
& & NP & $55$ & $27804$ \\
\cline{2-5}
& \multirow{3}{4em}{\textbf{SET3}} & Sparse & $112$ & $52896$ \\
& & WP & $112$ & $26448$ \\
& & NP & $83$ & $40482$ \\
\hline
\multirow{10}{4em}{\textbf{NET2}} & \multirow{1}{4em}{} & Baseline & $224$ & $110912$ \\
\cline{2-5}
& \multirow{3}{4em}{\textbf{SET1}} & Sparse & $224$ & $110912$ \\
& & WP & $224$ & $11091$ \\
& & NP & $44$ & $19256$ \\
\cline{2-5}
& \multirow{3}{4em}{\textbf{SET2}} & Sparse & $224$ & $110912$ \\
& & WP & $224$ & $33274$ \\
& & NP & $111$ & $48514$ \\
\cline{2-5}
& \multirow{3}{4em}{\textbf{SET3}} & Sparse & $224$ & $110912$ \\
& & WP & $224$ & $55456$ \\
& & NP & $167$ & $75662$ \\
\hline
\end{tabular}
\caption{Number of hidden neurons and of weights of the networks of interest. These values are valid for both the MNIST and FMNIST datasets. For the meaning of columns \textbf{Baseline}, \textbf{Param} and \textbf{Network} see Table \ref{tab:MNIST-acc-rob}. \textbf{\# HL Neurons} represent the number of neurons of the hidden layers of the network, \textbf{\# Weights} represent the number of connection in the network.}
\label{tab:nnw}
\end{table}

\paragraph{Accuracy and robustness} Tables \ref{tab:MNIST-acc-rob} and \ref{tab:FMNIST-acc-rob} show
the results of our analysis about the accuracy and the
robustness of the networks trained on MNIST and FMNIST,
respectively. The purpose of these data is to show that our training pipeline does
robustness and accuracy of our networks do not differ from baseline architectures in a
substantial way, which makes the approach of running verification to
certify pruned networks useful in practice. Should our networks
result in unsatisfactory accuracy or robustness rates, it would not be
interesting to subject them to further analysis. As it can be
observed, both the weight-pruned networks and the neuron-pruned ones
present a slight decrease in accuracy which correlates positively with
the strength of pruning, i.e., the larger the sparsity rate, the
higher the loss in accuracy. The results related to robustness are
less clear cut: on MNIST, robustness decreases in neuron
pruned networks, but it increases in weight pruned ones; on FMNIST
dataset the opposite appears to be true. Given this discrepancy, we
believe that further investigation on pruning techniques --- which is
beyond the scope of this paper --- might be in order. Here, we just
mention that decrease in robustness has been linked to
\textit{over-pruning} in~\cite{DBLP:conf/nips/GuoZZC18}, where networks are
considered to be  over-pruned whenever they are more than 50 times
smaller than their baseline; still in~\cite{DBLP:conf/nips/GuoZZC18},
more modest levels of pruning are linked with an increase in
robustness. To quantify the impact of our procedures on the resulting architectures, we report the number of 
neurons and weights surviving in the hidden layers of the pruned networks in Table~\ref{tab:nnw}.
As it can be observed, networks pruned with NP present a great number
of pruned connections besides the expected pruned neurons, but this
still does not explain the discrepancy observed in robustness.
Another explanation could be that the fine-tuning performed in
our pruning procedures is not oriented to achieve robustness --- which could be done using
techniques such as robust training~\cite{DBLP:conf/nips/WongSMK18}.
In this case, the robustness of the resulting network could be 
improved by using the proper training procedures and the overall results
could be made more predictable than our current setting.

\begin{table*}
\begin{tabular}{cc}
\begin{minipage}{0.47\textwidth}
\scalebox{0.86}{
\begin{tabular}{|c|c|c|c|c|c|}
\hline
\multicolumn{6}{|c|}{\textbf{MNIST}}\\
\hline
\textbf{Base} & \textbf{Param} & \textbf{Network} & \textbf{Marabou} & \textbf{MIPVerify} & \textbf{ERAN}\\
\hline
\multirow{7}{4em}{\emph{NET1}} & \multirow{1}{4em}{} & Baseline & $0$ & $0$ & $0$\\
\cline{2-6}
& \multirow{3}{4em}{\textbf{SET1}} & Sparse & $0*$ & $15*$ & $20$\\
& & WP & $8$ & $5*$ & $14$\\
& & NP & $18$ & $16*$ & $20$ \\
\cline{2-6}
& \multirow{3}{4em}{\textbf{SET2}} & Sparse & $0$ & $0$ & $1$\\
& & WP & $0$ & $0$ & $7$\\
& & NP & $5$ & $15$ & $17$ \\
\cline{2-6}
& \multirow{3}{4em}{\textbf{SET3}} & Sparse & $0$ & $0$ & $1$\\
& & WP & $0$ & $0$ & $1$\\
& & NP & $6$ & $0$ & $7$\\
\hline
\multirow{7}{4em}{\emph{NET2}} & \multirow{1}{4em}{} & Baseline & $0$ & $0$ & $0$\\
\cline{2-6}
& \multirow{3}{4em}{\textbf{SET1}} & Sparse & $0*$ & $17*$ & $20$\\
& & WP & $9$ & $0$ & $0$\\
& & NP & $18$ & $14*$ & $19$ \\
\cline{2-6}
& \multirow{3}{4em}{\textbf{SET2}} & Sparse & $0*$ & $0$ & $0$\\
& & WP & $0$ & $0$ & $0$\\
& & NP & $0$ & $0$ & $1$\\
\cline{2-6}
& \multirow{3}{4em}{\textbf{SET3}} & Sparse & $0$ & $0$ & $0$\\
& & WP & $0$ & $0$ & $0$\\
& & NP & $1$ & $0$ & $0$\\
\hline
\end{tabular}}
\end{minipage}
&
\begin{minipage}{0.47\textwidth}
\scalebox{0.86}{
\begin{tabular}{|c|c|c|c|c|c|}
\hline
\multicolumn{6}{|c|}{\textbf{FMNIST}}\\
\hline
\textbf{Base} & \textbf{Param} & \textbf{Network} & \textbf{Marabou} & \textbf{MIPVerify} & \textbf{ERAN}\\
\hline
\multirow{7}{4em}{\emph{NET1}} & \multirow{1}{4em}{} & Baseline & $0$ & $0$ & $1$\\
\cline{2-6}
& \multirow{3}{4em}{\textbf{SET1}} & Sparse & $0*$ & $20$ & $20$\\
& & WP & $2$ & $5$ & $19$\\
& & NP & $19$ & $20$ & $20$ \\
\cline{2-6}
& \multirow{3}{4em}{\textbf{SET2}} & Sparse & $0$ & $0$ & $0$\\
& & WP & $0$ & $0$ & $5$\\
& & NP & $6$ & $4$ & $20$ \\
\cline{2-6}
& \multirow{3}{4em}{\textbf{SET3}} & Sparse & $0$ & $0$ & $0$\\
& & WP & $3$ & $0$ & $3$\\
& & NP & $0$ & $0$ & $4$\\
\hline
\multirow{7}{4em}{\emph{NET2}} & \multirow{1}{4em}{} & Baseline & $0$ & $0$ & $1$\\
\cline{2-6}
& \multirow{3}{4em}{\textbf{SET1}} & Sparse & $0*$ & $19*$ & $20$\\
& & WP & $0$ & $0$ & $0$\\
& & NP & $0*$ & $17$ & $20$ \\
\cline{2-6}
& \multirow{3}{4em}{\textbf{SET2}} & Sparse & $0*$ & $0$ & $0$\\
& & WP & $0$ & $0$ & $0$\\
& & NP & $0$ & $0$ & $0$ \\
\cline{2-6}
& \multirow{3}{4em}{\textbf{SET3}} & Sparse & $0*$ & $0$ & $0$\\
& & WP & $0$ & $0$ & $0$\\
& & NP & $0$ & $0$ & $0$ \\
\hline
\end{tabular}}
\end{minipage}
\end{tabular}
\caption{Results of our experiment for Marabou, ERAN and
  MIPVerify. The values reported represent the number of problems
  which were solved successfully within the timeout. See Table
  \ref{tab:FMNIST-acc-rob} for the meaning of the columns
  \textbf{Param}, \textbf{Network} and \textbf{Base}.
  \textbf{Marabou}, \textbf{MIPVerify} and \textbf{ERAN}
  represent the number of problems solved by Marabou, MIPVerify and
  ERAN, respectively.}
\label{tab:ver-result}
\end{table*}

\paragraph{Impact on verification} Table \ref{tab:ver-result} reports the results of our experiments
with the verification tools we consider.
In particular we represent the number of problems, i.e.,
adversarial searches, that each tool managed to complete successfully
with respect to the different datasets, sets of parameters and sets of
images. It should be noted that the number of solvable problems is
always 20, except for Marabou, where
the number of problems is 19. 
As it can be observed, the general trend for all the tools is
that neuron-pruned networks are verified more easily than other
configurations, in particular when the pruning is made more aggressive.
However, there are some exceptions to this trend. For instance, we could observe that when NET2 networks are strongly
pruned (\textit{i.e.}, NET2 and SET1), MIPVerify is able to solve more
instances on the sparse networks than on the neuron pruned ones.

Weight pruning appears to be consistently less effective than neuron pruning and,
in the case of strongly pruned NET2 networks, even than the sparse training.
It should be noted that for the SET1 level of pruning Marabou and MIPVerify present
unexpected behaviours: in some cases Marabou terminates immediately reporting that the verification query was not satisfiable,
whereas MIPVerify returns an array of Not a Number. The experiments in which the tools
were subject to the above mentioned problem are identified with a $*$ beside the number
of instances verified (the instances in which the problem arised were considered as timed out). We believe these behaviours might be the result of numerical instabilities caused by matrix sparsity. In the case of MIPVerify, this hypothesis has been confirmed by its developers. Gurobi, the underlying solver used by MIPVerify, is affected by the bad conditioning of the sparse matrices our procedures produce. 

In Table~\ref{tab:ver-result} we do not report
runtimes on the solved instances, but if we consider them then we see
that Marabou and MIPVerify have comparable performances:
Marabou is better than MIpVerify on FMNIST but it is worse on
MNIST. On the other hand, ERAN seems to be consistently faster than
Marabou and MIPVerify. As we expected, all the tools appear to be, on
average, faster on networks which are pruned more aggressively.

\paragraph{Effects of WP and NP on complete methods} We now turn our attention to the different effects NP and WP had on the performances of verification tools. We postulate that the observed differences can be explained by considering the techniques at the core of the
verification algorithms considered. Indeed, all three tools resort to
constraint-based methodologies (SMT and MILP) whose performances are
sensitive to the number of variables and constraints to be found in
the encoding of the original problem. The compilation of the
non-linear activation functions requires a (potentially large) number
of variables and constraints to represent them, and thus reducing
neurons has a direct impact on the sheer complexity of the encoding.
On the other hand, weight pruning eliminates mostly arithmetical
operations but leaves the number of variables and constraints mostly
unchanged. To support these hypothesis we have performed some
additional experimental analysis on the version of ERAN relying on
abstract interpretation only --- the results are presented in
Table~\ref{tab:res-eran-inc}. In this ``incomplete'' version, ERAN does not  
make use of a MILP solver in order to find the adversarial example, but
it just analyses the robustness of the network through sound
overapproximations. ERAN-incomplete is able to manage
networks more efficiently than Marabou and MIPVerifiy, but abstract
adversarial examples may no longer succeed to map to concrete ones. 
As it can be seen in Table~\ref{tab:res-eran-inc}, ERAN-incomplete
manages to verify weight-pruned networks in less time than
neuron-pruned ones, and the speed-up appears to be directly related to
the number of weights pruned.

Regarding the results obtained for the sparse networks when heavily pruned,
we believe that MIPVerify and ERAN manage to leverage the presence of many
connection weights set to zero in order to remove non-linearity from their MILP problems.
In particular whenever a non-linear activation function presents only zero values
connections in input or in output it is removed, since it would not influence
the network in any way. To confirm our hypothesis we have analysed the log files
of Gurobi when called by MIPVerify on the weight pruned, neuron pruned and sparse networks: as we
expected the problem solved by Gurobi in the case of neuron pruned and sparse networks present
less than $1/4$ of integer variables with respect to the problem solved on the weight pruned networks.

\begin{table}
\centering
\begin{tabular}{|c|c|c|c|c|}
\hline
\multicolumn{5}{|c|}{\textbf{ERAN - incomplete}}\\
\hline
\textbf{Base} & \textbf{Param} & \textbf{Network} & \textbf{MNIST} & \textbf{FMNIST}\\
\hline
\multirow{7}{4em}{\emph{NET1}} & \multirow{1}{4em}{} & Baseline & $2.12s$ & $2.15s$ \\
\cline{2-5}
& \multirow{3}{4em}{\textbf{SET1}} & Sparse & $1.38s$ & $1.37s$ \\
& & WP & $0.23s$ & $0.27s$ \\
& & NP & $0.40s$ & $0.25s$ \\
\cline{2-5}
& \multirow{3}{4em}{\textbf{SET2}} & Sparse & $2.15s$ & $2.17s$ \\
& & WP & $0.68s$ & $0.77s$ \\
& & NP & $1.02s$ & $0.94s$ \\
\cline{2-5}
& \multirow{3}{4em}{\textbf{SET3}} & Sparse & $2.16s$ & $2.18s$ \\
& & WP & $1.10s$ & $1.15s$ \\
& & NP & $1.56s$ & $1.58s$ \\
\hline
\multirow{7}{4em}{\emph{NET2}} & \multirow{1}{4em}{} & Baseline & $6.52$ & $6.60$ \\
\cline{2-5}
& \multirow{3}{4em}{\textbf{SET1}} & Sparse & $5.48s$ & $5.45s$ \\
& & WP & $0.88s$ & $1.18s$ \\
& & NP & $1.01s$ & $0.74s$ \\
\cline{2-5}
& \multirow{3}{4em}{\textbf{SET2}} & Sparse & $6.24s$ & $6.25s$ \\
& & WP & $2.15s$ & $2.37s$ \\
& & NP & $2.33s$ & $2.24s$ \\
\cline{2-5}
& \multirow{3}{4em}{\textbf{SET3}} & Sparse & $5.96s$ & $5.86s$ \\
& & WP & $3.19s$ & $3.38s$ \\
& & NP & $3.89s$ & $1.28s$ \\
\hline
\end{tabular}
\caption{Results of our experiment for the datasets MNIST and FMNIST on the incomplete version of ERAN. See Table \ref{tab:FMNIST-acc-rob} for the meaning of the columns \textbf{Param} and \textbf{Network}. The columns MNIST and FMNIST report the average times needed to solve a single adversarial search for MNIST and FMNIST datasets, respectively. The averages are computed on 30 different images.}
\label{tab:res-eran-inc}
\end{table}

%% file: conclusions.tex
\section{Conclusions and Future Works}\label{sec:conclusions}
One of the main challenges with verification of (deep) neural networks is
the scalability of the currently available tools. While a great deal
of research effort has been put in developing new and more scalable 
verification tools, leveraging statistical methods from the machine
learning community to ease formal analysis has been mostly
ignored. In this paper we have provided evidence that the interaction
between pruning methods and verification tools can be effective
and enable formal analysis of networks that could not be checked
otherwise. We have studied how two different pruning  methods can be
embedded in a training pipeline to produce networks with good
accuracy and robustness, while lowering the complexity of their corresponding verification problem. 
As future research, we plan to extend the work presented to convolutional
neural networks, considering also other pruning methods, e.g., filter
pruning, and verification tools. Moreover we intend to investigate the
robustness property of the pruned networks, both using different
pruning methods and testing them using robust training and other
analogous techniques.